# Constrained Deep Transfer Feature Learning and its Applications

Yue Wu    Qiang Ji
ECSE Department, Rensselaer Polytechnic Institute
110 8th street, Troy, NY, USA
{wuy9, jiq}@rpi.edu

## Abstract

*Feature learning with deep models has achieved impressive results for both data representation and classification for various vision tasks. Deep feature learning, however, typically requires a large amount of training data, which may not be feasible for some application domains. Transfer learning can be one of the approaches to alleviate this problem by transferring data from data-rich source domain to data-scarce target domain. Existing transfer learning methods typically perform one-shot transfer learning and often ignore the specific properties that the transferred data must satisfy. To address these issues, we introduce a constrained deep transfer feature learning method to perform simultaneous transfer learning and feature learning by performing transfer learning in a progressively improving feature space iteratively in order to better narrow the gap between the target domain and the source domain for effective transfer of the data from source domain to target domain. Furthermore, we propose to exploit the target domain knowledge and incorporate such prior knowledge as constraint during transfer learning to ensure that the transferred data satisfies certain properties of the target domain.*

*To demonstrate the effectiveness of the proposed constrained deep transfer feature learning method, we apply it to thermal feature learning for eye detection by transferring from the visible domain. We also applied the proposed method for cross-view facial expression recognition as a second application. The experimental results demonstrate the effectiveness of the proposed method for both applications.*

## 1. Introduction

Feature learning with deep models is an active research area. Recent research has demonstrated that with feature learning methods, effective features can be learnt for both representation and classification of the input data for many computer vision tasks including face recognition [20], object detection [5], and scene classification [24]. Feature learning with deep model, however, typically requires a large amount of training data. Hence, feature learning for domains with scarce data is not feasible.

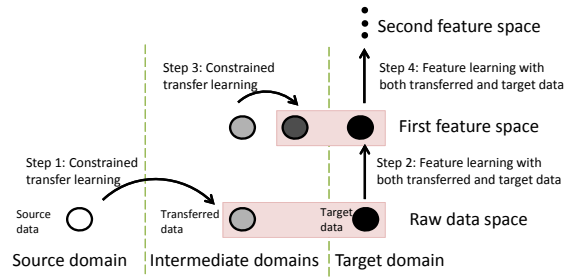

Figure 1. The framework of the proposed constrained deep transfer feature learning method. The algorithm iteratively performs transfer learning and feature learning in more and more deeper feature spaces.

Transfer learning can be one of the approaches to address this problem and help feature learning in the data-scarce target domain by transferring data or knowledge from data-rich source domain. Transfer learning not only can compensate for the lack of data in the target domain but can also benefit the tasks in the target domain from the experience gained from the source domain. The transfer learning techniques usually involve instance transfer, feature transfer, model parameter transfer, or relational knowledge transfer [12]. Those transfer learning techniques have been used in natural language processing [1], document classification [23], etc. However, typical transfer learning techniques usually perform one-shot transfer in a fixed or shallow feature space, while a fixed feature space may not effectively fill the semantic gap between the target and source domains. Another issue with transfer learning is that it is purely data driven, without adequately considering certain inherent properties of the target data.

To tackle these issues, we propose a constrained deep transfer feature learning method to perform simultaneous transfer learning and feature learning by exploiting the knowledge in both target and source domains. The general framework is shown in Figure 1. Specifically, we propose to iteratively perform constrained transfer learning and feature learning in several increasingly deeper feature spaces to gradually transfer the knowledge from source domain to target domain and learn the features in the target domain. Furthermore, we propose to exploit the target domain knowledge and incorporate such prior knowledge as constraint during transfer learning to ensure that the transferred data satisfies certain properties of the target domain principles.

The proposed framework has the following merits. First, the progressive transfer allows the creation of several intermediate pseudo domains to bridge the gap between the source and target domains. As the knowledge transfer continues, the intermediate domains gradually approach the target domain. Second, feature learning is performed at each level as knowledge transfer happens, and it is also performed progressively at a higher level as knowledge transfer continues. Hence, knowledge transfer and feature learning intertwine at each step, improving both feature learning and knowledge transfer. Finally, by imposing constraints on the transferred data, we can ensure the transferred data not only possess certain desired characteristics but also to be semantically meaningful.

The remaining part of this paper is organized as follows. Section 2 reviews the related work. Section 3 discusses the Restricted Boltzmann Machine and Deep Boltzmann Machine models. Section 4 introduces the proposed method. Section 5 shows the experimental results. Section 6 concludes the paper.

## 2. Related Work

Recently, an increasing number of works concentrate on learning good representations for data with deep models. In [18], Salakhutdinov et al. build a hierarchical deep model to learn features for object recognition and handwritten character recognition. Similar to the proposed method, some works perform feature learning based on multimodal data in different domains. For example, in [11], Ngiam et al. use deep autoencoder to learn common features from audio and video data. In [19], a deep multimodal DBM is constructed to learn shared features for images and texts. However, in contrast to the existing works that learn shared representations across domains [11][19], our work focuses on transferring knowledge from one domain to help feature learning for another domain. Multimodal feature learning methods usually require a lot of paired training data, while we specifically handle the case where there is limited data in the target domain and rich data in the source domain.

Transfer learning refers to the learning methods that leverage the knowledge from the source domain to help learning in the target domain. The transferred knowledge includes training instances, features, model parameters and relational knowledge [12]. For example, in [3], Triadaboost is proposed to transfer the instances within the framework

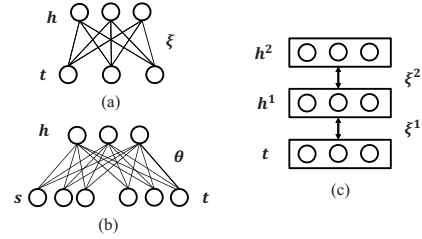

Figure 2. (a) RBM for data in one domain. (b) RBM for transfer learning. (c) DBM model.

of adaboost. In [15], feature spaces are found that minimize the inter-domain differences.

There is also work that combines transfer learning with feature learning. In [23], deep features learned with the DBM model using data in the source domain are selected for document classification in the target domain. Different from the work [23] that first learns deep features and then transfers the learned features among multimodal data, our work preforms transfer feature learning at multiple levels of the deep architecture. More importantly, we add target domain knowledge to constrain the transfer learning and feature learning procedure.

## 3. Background

Before we introduce the proposed method, we first review the Restricted Boltzmann Machine (RBM) and the Deep Boltzmann Machine (DBM) models. RBM is a undirected probabilistic graphical model (Figure 2 (a)) that captures the joint probability of the binary input data $\mathbf{t}$ (e.g. data in the target domain) with multiple binary hidden nodes $\mathbf{h}$.

$$p(\mathbf{t}; \xi) = \frac{\sum_{\mathbf{h}} exp(-E(\mathbf{t}, \mathbf{h}; \xi))}{Z(\xi)}, \quad (1)$$

$$-E(\mathbf{t}, \mathbf{h}; \xi) = \mathbf{t}^T W \mathbf{h} + \mathbf{c}^T \mathbf{t} + \mathbf{b}^T \mathbf{h}, \quad (2)$$

where $E(.)$ is the energy function, $Z(\xi) = \sum_{\mathbf{t},\mathbf{h}} exp(-E(\mathbf{t}, \mathbf{h}; \xi))$ is the partition function, and $\xi = \{W, \mathbf{c}, \mathbf{b}\}$ are the parameters. The conditional probabilities are as follows:

$$p(t_i = 1|\mathbf{h}; \xi) = \sigma(\sum_j W_{i,j} h_j + c_i), \quad (3)$$

$$p(h_j = 1|\mathbf{t}; \xi) = \sigma(\sum_i t_i W_{i,j} + b_j), \quad (4)$$

where $\sigma(.)$ denotes the sigmoid function. Given the training data, the parameters are learned by maximizing the log likelihood with stochastic gradient ascend algorithm approximated by the Contrastive Divergence (CD) algorithm [7].

DBM model [16] shown in Figure 2 (c) (assume two layer model thereafter) consists of one layer of visible nodes and multiple layers of binary hidden nodes. It represents the probability of visible nodes (similar as Equation 1) with the new energy function (ignoring the bias terms):

$$-E(\mathbf{t}, \mathbf{h}^1, \mathbf{h}^2; \xi^1, \xi^2) = \mathbf{t}^T W^1 \mathbf{h}^1 + (\mathbf{h}^1)^T W^2 \mathbf{h}^2, \quad (5)$$

Table 1. Notations. **t**, **s** and $\widetilde{\mathbf{t}}$ represent target data, source data, and the pseudo target data transferred from the source domain. If necessary, we use the superscript to indicate the pairwise (P)/unpaired (U) data (corresponding data in both domains, and data in one domain).

| Data | Notations |
|---|---|
| Pairwise target and source data | $Data^P = \{\mathbf{t}_k, \mathbf{s}_k\}_{k=1}^{N^P}$ |
| Unpaired target data | $Data_t^U = \{\mathbf{t}_i\}_{i=1}^{N_t^U}$ |
| Unpaired source data | $Data_s^U = \{\mathbf{s}_j\}_{j=1}^{N_s^U}$ |
| Unpaired pseudo target data | $Data_{\tilde{t}}^U = \{\widetilde{\mathbf{t}}_n\}_{n=1}^{N_{\tilde{t}}^U}$ |

where $\xi = \{\xi^1, \xi^2\}$ are the parameters for different layers. DBM is learned in a layer-wise manner and then jointly fine tuned [16].

For feature learning using RBM or DBM, the hidden nodes in the top layer are inferred for each input data using Equation 4 or the mean-field techniques [16], and they are considered as the new feature representations. For transfer learning, the input data is the concatenated data from the source **s** and target domains **t**, and the RBM (Figure 2 (b)) will learn their joint probability distribution. Transfer learning can then be performed using the joint probability.

## 4. Constrained deep transfer feature learning

### 4.1. The general framework

The proposed constrained deep transfer feature learning method is motivated by the following intuitions. First, it's straightforward to think of directly applying the DBM model illustrated in section 3 to learn the deep features for the target domain. However, DBM learning usually requires a lot of training data with variations, while there is limited data in the target domain. But, if we can capture the joint distribution of target and source data as a bridge between the source and target domains, we could transfer additional large amount of source data to the target domain as pseudo data for further target domain feature learning. Second, because of the significant domain differences, one time transfer may not be effective as stated above. Therefore, we propose to perform the transfer feature learning progressively at different levels of feature spaces to generate intermediate pseudo target spaces to gradually approach the final target space. Third, because of the underlying mechanism that produces the data, target data must follow certain properties that we want to preserve in the transferred data. We hence constrain the transfer learning to ensure the satisfaction of these properties. Therefore, we propose the constrained deep transfer feature learning method. Table 1 shows the notations.

The general framework is shown in Figure 1 and Algorithm 1. First, in the pre-training stage, we perform the transfer learning and feature learning iteratively. In the transfer learning step, we capture the joint distribution of

**Algorithm 1**: Constrained deep transfer feature learning

**Data**: Pairwise source and target data ($Data^P$); Unpaired target data ($Data_t^U$); Unpaired source data ($Data_s^U$).
**Result**: Learned features: $\xi^1, \xi^2,...$
/* Pre-training stage                              */
**for** *Layer l=1 to L* **do**
  /* Constrained transfer learning                 */
  • Learn the joint probability of target and source data $p(\mathbf{t}^l, \mathbf{s}^l; \theta^l)$ with constraint $\mathbf{C}(.; \theta^l)$.
  • Transfer the unpaired source data $\mathbf{s}_i^l$ by sampling the pseudo target data $\widetilde{\mathbf{t}}^l$ through $p(\mathbf{t}^l|\mathbf{s}_i^l; \theta^l)$.
  /* Feature learning                              */
  • Learn the features $\xi^l$ with target and pseudo target data.
  • Project the data to the learned feature space for further constrained transfer feature learning.
**end**
/* Joint fine-tuning stage                         */
  • Jointly fine-tune the features $\xi^1, \xi^2, ...$ based on the deeply transferred pseudo target data and the real target data.

target and source data $p(\mathbf{t}, \mathbf{s}; \theta)$ using RBM (Figure 2 (b)) as a bridge and transfer additional large amount of source data as pseudo target data $\widetilde{\mathbf{t}}$ by sampling through the conditional distribution $p(\mathbf{t}|\mathbf{s}; \theta)$. In addition, we impose constraint $\mathbf{C}(.; \theta)$ in the learning. Third, in the feature learning step, the transferred pseudo target data are combined with the original target data (shaded area in Figure 1) for feature learning using RBM (Figure 2 (a)). Then, the pseudo target data and real target data are projected to the learned feature space. In the new feature space of the next level, the transferred pseudo target data are considered as "source data" for further constrained transfer feature learning in the next iteration. Finally, similar as the DBM model (Figure 2 (c)), the layer-wisely learned features with parameters $\xi^1, \xi^2,...$ are fine-tuned jointly based on the deeply transferred pseudo target data and the real target data. The iterative transfer learning and feature learning should converge, because deep feature learning has been shown converging [8] and the transfer learning also converges due to the gradually reduced gaps between target domain and source domain through the iterations. In the following subsections, we discuss each step in details.

### 4.2. Constrained semi-supervised transfer learning in one layer

In this section, we first discuss how to learn the joint probability of target and source data $p(\mathbf{t}, \mathbf{s}; \theta)$ in a semi-supervised manner with constraints. Then, we discuss how to generate the pseudo target data by sampling through $p(\mathbf{t}|\mathbf{s}; \theta)$.

### 4.2.1 Semi-supervised learning of the joint probability

In this work, we use RBM model illustrated in section 3 to learn the joint probability of target and source data $p(\mathbf{t}, \mathbf{s}; \theta)$. As shown in Figure 2 (b), the model defines the joint probability:

$$p(\mathbf{t}, \mathbf{s}; \theta) = \frac{\sum_\mathbf{h} exp(-E(\mathbf{t}, \mathbf{s}, \mathbf{h}; \theta))}{Z(\theta)}, \quad (6)$$

where the energy function $E(.)$ has similar format as that in Equation 2.

Recall that RBM is usually learned with the Contrastive Divergence (CD) algorithm [7]. However, standard CD learning for the RBM model requires a large amount of pairwise source and target data, which are limited in our application. To tackle this problem, we propose the semi-supervised learning method. Specifically, we learn the model parameters by maximizing the log likelihood w.r.t the pairwise source and target data $Data^P$, the unpaired target data $Data_t^U$, and the unpaired source data $Data_s^U$.

$$\theta^* = arg \max_\theta L(\theta; Data) \quad (7)$$

$$L(\theta; Data) = L(\theta; Data^P) + \alpha L(\theta; Data_t^U) + \beta L(\theta; Data_s^U) \quad (8)$$

$$L(\theta; Data^P) = \frac{1}{N^P} \sum_{k=1}^{N^P} log(p(\mathbf{t}_k, \mathbf{s}_k; \theta)) \quad (9)$$

$$L(\theta; Data_t^U) = \frac{1}{N_t^U} \sum_{i=1}^{N_t^U} log(p(\mathbf{t}_i; \theta)) \quad (10)$$

$$L(\theta; Data_s^U) = \frac{1}{N_s^U} \sum_{j=1}^{N_s^U} log(p(\mathbf{s}_j; \theta)) \quad (11)$$

Here, $\alpha$ and $\beta$ are two parameters that balance different terms. $p(\mathbf{t}; \theta)$ and $p(\mathbf{s}; \theta)$ in Equation 10 and Equation 11 are the marginal distributions of data in one domain by summing out the missing data in the other domain.

Following the CD algorithm, we solve the optimization problem in Equation 7 using gradient ascent algorithm. The gradient of model parameters for each log likelihood term is calculated as:

$$\frac{\partial L(\theta; Data)}{\partial \theta} = -\langle \frac{\partial E}{\partial \theta} \rangle_{P_{data}} + \langle \frac{\partial E}{\partial \theta} \rangle_{P_{model}}, \quad (12)$$

where $\langle . \rangle_{P_{data}}$ and $\langle . \rangle_{P_{model}}$ represent the data dependent and model dependent expectations.

In the CD algorithm [7], to approximately calculate the data dependent expectation $\langle . \rangle_{P_{data}}$, we need to sample the unknown variables from the data dependent probabilities $P_{data}$. Note that, in the semi-supervised learning setting, the data dependent probabilities $P_{data}$ differ for each data type. Specifically, $P_{data^P} = p(\mathbf{h}|\mathbf{t}_k, \mathbf{s}_k; \theta)$, $P_{data_t^U} = p(\mathbf{h}, \mathbf{s}|\mathbf{t}_i; \theta)$ and $P_{data_s^U} = p(\mathbf{h}, \mathbf{t}|\mathbf{s}_j; \theta)$. For the pairwise data, we could directly sample $\mathbf{h}$ from $p(\mathbf{h}|\mathbf{t}_k, \mathbf{s}_k; \theta)$ using Equation 4. However, for the unpaired data, $p(\mathbf{h}, \mathbf{s}|\mathbf{t}_i; \theta)$ and $p(\mathbf{h}, \mathbf{t}|\mathbf{s}_j; \theta)$ are intractable. Thus, we use Gibbs sampling to generate $\mathbf{h}, \mathbf{s}$ from $p(\mathbf{h}, \mathbf{s}|\mathbf{t}_i; \theta)$. Similarly, we can generate samples $\mathbf{h}, \mathbf{t}$ from $p(\mathbf{h}, \mathbf{t}|\mathbf{s}_j; \theta)$. Following the CD algorithm, we estimate the model dependent expectations with k-step (k=5) Gibbs update through the model, starting from the samples calculated using the data dependent probabilities.

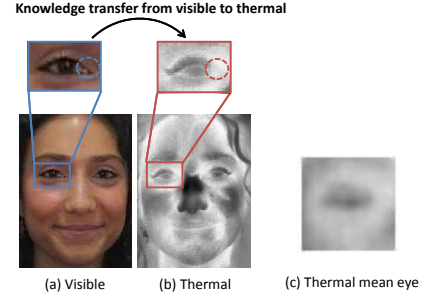

Figure 3. Constrained deep transfer feature learning for thermal eye detection. (a)(b) Pairwise visible and thermal facial images [13]. (c) Thermal mean eye.

### 4.2.2 Transfer learning with constraints

In every domain, its data only represent the observations of the underlying latent objects. It is often through either a physical or biological process that the latent objects produce the observed data in the target domain. For example, as shown in Figure 3, in the case of eye detection on thermal images by transferring the knowledge from visible domain, the intensities of the eyes measure the temperature near the eye skin surface and the eye temperature is determined by the the blood flow to the eye skin surface. Hence, the intensity distribution near the eye on an thermal image is mainly determined by the underlying latent vascular structure distribution. For example, the tear dust area usually has the high temperature due to the blood flow in the artery beneath (Figure 3 (c)). On the other hand, in the pupil area and the hair-insulated eye lash region, the temperature is expected to be low because of lack of blood flow. These unique facial anatomy structures in the eye regions lead to the unique and distinct target data pattern that is universal across subjects. As a result, to ensure a physically and semantically meaningful transfer, we propose to impose certain constraints during the transfer learning in order to preserve such target data properties, such as the certain unique shape and appearance characteristics.

Based on the intuitions illustrated above, we modify the parameter learning problem in Equation 7 by adding the constraint $\mathbf{C}(.; \theta)$:

$$\theta^* = arg \max_\theta L(\theta; Data) - \lambda \mathbf{C}(.; \theta) \quad (13)$$

Here, the first term is the log likelihood function defined in Equation 8. The second term $\mathbf{C}(.; \theta)$ is a cost function to ensure that the transferred data satisfies certain properties.

To solve this optimization problem in Equation 13, depending on the type of $\mathbf{C}(.;\theta)$, we could still use the gradient ascent algorithm and we have derived the gradient of the first term w.r.t the parameters in section 4.2.1. Then, we only need to calculate the gradient of the second term $\mathbf{C}(.;\theta)$ w.r.t the parameters, i.e., $\frac{d\mathbf{C}(.;\theta)}{d\theta}$. It's detailed calculated will be discussed in section 5.2.

#### 4.2.3 Pseudo target data generation by sampling

Given the learned joint probability $p(\mathbf{t},\mathbf{s};\theta)$, we could transfer the large amount of additional source data $\mathbf{s}_j \in Data_s^U$ to the target domain to generate the pseudo target data $\widetilde{\mathbf{t}}$. This can be done by sampling through $p(\mathbf{t}|\mathbf{s}_j;\theta), \mathbf{s}_j \in Data_s^U$ using Gibbs Sampling method that iteratively calls Equation 3 and 4.

### 4.3. Feature learning in one layer

In each iteration, combining the transferred pseudo target data and the real target data, the goal of feature learning is to learn the RBM model (Figure 2 (a)) that captures the variations of target and pseudo target data and uses the hidden nodes as new features. More formally, the RBM is defined in Equation 1 and parameter learning is formulated as:

$$\xi^* = arg \max_{\xi} L(\xi; Data_t) + \gamma L(\xi; Data_{\widetilde{t}}^U) \quad (14)$$

The first term and second term represent the log likelihood w.r.t the real target data $Data_t = Data_t^P \bigcup Data_t^U$, and the pseudo target data $Data_{\widetilde{t}}^U$. $\gamma$ is the parameter that balances the two terms. To learn the features, we apply the standard Contrastive Divergence (CD) algorithm [7]. The only difference is that the gradient is calculated based on both terms in Equation 14. Given the learned model, for each target and pseudo target data, its new representation can be calculated using Equation 4. Then, for the next iteration in the learned feature space, constrained transfer learning and feature learning continue and interact until convergence.

### 4.4. Joint feature fine-tuning

After multi-layer constrained transfer feature learning in the pre-training stage, we have the initially learned features $\xi^1, \xi^2, ...$. In addition, we have the deeply transferred pseudo target data in the top feature layer denoted as $\widetilde{\mathbf{t}}^L$. Then, for joint feature fine-tuning, we need to project the pseudo data back to the original space by repeatedly calling Equation 3 with parameters $\xi^L,..., \xi^1$ and get the deeply transferred pseudo data in the original space $\widehat{\mathbf{t}}^1$. Then, based on the pseudo data $\widehat{\mathbf{t}}^1$ and the real target data $\mathbf{t}^1$, we apply standard fine-tuning algorithm [16] for the Deep Boltzmann Machine model, where the mean-field fix point equation is used to estimate the data-dependent expectation and the Persistent Markov Chain is used to estimate the model dependent expectation.

## 5. Experimental Results

### 5.1. Eye detection on thermal images by transferring from visible domain

To demonstrate the proposed framework for constrained deep transfer feature learning, we applied it to eye detection on thermal images. Comparing to facial analysis in the visible domain, thermal facial analysis is more sensitive to eye localizations [2]. However, there are limited works about eye detection on thermal images. Furthermore, the existing thermal eye detection techniques typically use the visible image features, which are suboptimal, since thermal and visible images are formed based on different principles [4]. For example, thermal images contain limited texture and gradient information due to the heat diffusion phenomenon, while the visible image features usually focus on encoding these detailed information.

While feature learning on thermal images can alleviate this problem, limited thermal training images make it very difficult to leverage on the existing deep learning models. To address this challenge, we propose to apply the proposed constrained transfer feature learning method to perform thermal feature learning by transferring the eye data in the visible domain to the thermal domain to compensate for the lack of thermal data. The target domain refers to the thermal patches, including the eye patches and the background patches. The source domain refers to the visible eye patches. Note that, we only need to transfer the eye patches as positive data, since we can generate many negative data by sampling from the background. For thermal eye detection, with the learned thermal features using the proposed method, we train SVM classifier to search the eye with a scanning window manner.

### 5.2. Implementation details

**Databases:** We use four databases including the visible and thermal facial behavior database(VTFB), the MAHNOB laughter database [13], the Natural visible and thermal facial expression database (NVIE) [21], and the Facial Recognition Technology (FERET) database [14]. VTFB has synchronized visible and thermal videos (FLIR SC6800 thermal camera) for 7 subjects and additional thermal videos for 13 subjects with spontaneous facial expressions and arbitrary head poses. The MAHNOB Laughter database provides thermal facial sequences of 22 subjects with moderate head poses, neutral and happy facial expressions. The NVIE database contains thermal facial sequences of 215 subjects with spontaneous and posed expressions. FERET database provides visible facial images with different head poses and moderate expressions.

During training, we use the synchronized thermal and visible images of the first 7 subjects from VTFB (1295 images) as pairwise data. We use the thermal images of ad-

5105

ditional 3 subjects from the VTFB (573 images), and visible images from FERET (3830 images) as unpaired data. We test the method on the thermal images of remaining 10 subjects from VTFB (542 images), the MAHNOB database (114 images) and the NVIE database (35550 images).

**The method:** We use two layer DBM with 800 and 600 hidden nodes to learn features. The RBM models used to learn the joint distributions $p(\mathbf{t}, \mathbf{s}; \theta^l)$ have 400 hidden nodes. The hyper-parameters $\alpha$, $\beta$, $\lambda$ and $\gamma$ are 0.4, 0.3, 0.002, and 0.3, respectively. In our experiments, the height and width of the image patches are about 1/2 the inter-ocular distance and the patches are normalized to 40×40. The background patches are at least 1/4 inter-ocular distance away from the eyes. We augment the training data by rotating and resizing the images. Following [16], we learn the Gaussian-binary RBMs [9] with 1000 hidden nodes for the raw visible and thermal image patches, respectively, and then treat the values of the hidden layers as the preprocessed data to speed up the learning procedure. With Matlab implementation, it takes about 30 hours to train the full model with two layers and the constraint on a single core machine.

To impose the target domain constraint as discussed in section 4.2.2, we propose to impose the constraint that the transferred thermal eyes must satisfy the general appearance pattern as shown in Fig. 3(c) for the thermal eye. To achieve this goal, we propose to capture such a target data pattern by using a thermal mean eye, which can be obtained by averaging the existing target data. The basic idea is that while individual thermal eye may vary because of each person's unique eye structure, through averaging, the thermal mean eye can capture their commonality while canceling out their differences. The commonality is resulted from the shared underlying eye structure. Figure 3(c) shows the mean eye as an example. It apparently can capture the unique pattern of a thermal eye, i.e., brighter in the inner eye corner and darker in the eye center that are true across subjects at different conditions.

For transferring the visible eyes to thermal domain to help the thermal feature learning, the transfer learning step in Equation 13 becomes:

$$\theta^* = \arg\max_\theta L(\theta; Data^+) - \lambda \|\frac{1}{N_s^{U+}} \sum_j \langle \mathbf{t}^+ \rangle_{Q_j} - \mathbf{m}^+\|_2^2. \quad (15)$$

Here, "+" denotes the positive eye data. The second term enforces the mean of the transferred pseudo thermal eye $\frac{1}{N_s^{U+}} \sum_j \langle \mathbf{t}^+ \rangle_{Q_j}$ is close to the given mean eye $\mathbf{m}^+$. Specifically, for each source data $\mathbf{s}_j^+ \in Data_s^{U+}$, $\langle \mathbf{t} \rangle_{Q_j}$ represents the expected transferred pseudo thermal eye and $Q_j = p(\mathbf{t}^+|\mathbf{s}_j^+; \theta)$. Then, $\frac{1}{N_s^{U+}} \sum_j \langle \mathbf{t}^+ \rangle_{Q_j}$ will be the mean of the transferred pseudo thermal eyes and it should be close to the given mean eye $\mathbf{m}^+$. Assume the gradient of the second constraint term of the objective function in Equation 15 w.r.t $\theta$ is denoted as $\delta$. Then, it is calculated as follows:

$$\begin{aligned}\delta =& 2 * [\frac{1}{N_s^{U+}} \sum_j \langle \mathbf{t}^+ \rangle_{Q_j} - \mathbf{m}^+]^T \\ & * \frac{1}{N_s^{U+}} \sum_j [-\langle \mathbf{t}^+ \frac{\partial E}{\partial \theta} \rangle_{R_j} + \langle \frac{\partial E}{\partial \theta} \rangle_{R_j} \langle \mathbf{t}^+ \rangle_{Q_j}],\end{aligned} \quad (16)$$

where $R_j = p(\mathbf{t}^+, \mathbf{h}|\mathbf{s}_j^+; \theta), \mathbf{s}_j^+ \in Data_s^{U+}$.

**Evaluation criterion:** The eye detection error is defined as $error = \frac{max(\|D_l - G_l\|_2, \|D_r - G_r\|_2)}{\|G_l - G_r\|_2}$, where $D$ and $G$ represent the detected and ground truth eye locations, and the subscript denotes left and right eyes. We regard $error < 0.15$ as the successful detection. The evaluation criteria follows the methods [10] [22] for fair comparison.

### 5.3. Evaluation of the proposed method

In this section, we evaluate the proposed constrained deep transfer feature learning method on the VTFB database. Note that, in the constrained transfer learning step (discussed in section 4.2), the joint probability of thermal and visible eyes $p(\mathbf{t}^+, \mathbf{s}^+; \theta)$ can be learned with different approaches. They are (**M**1) learning with pairwise data using standard CD algorithm [7], (**M**2) proposed semi-supervised learning with paired and unpaired data as discussed in section 4.2.1, (**M**3) learning with pairwise data and constraints (second term in Equation 13), and (**M**4) learning with semi-supervised data and constraints (the full model). In the experiments, we compare the four variations and study the other properties of the proposed method.

#### 5.3.1 Evaluation of the constrained transfer learning method in layer one

First, we evaluate the constrained transfer learning method in layer one. We evaluate the learned joint probabilities $p(\mathbf{t}^+, \mathbf{s}^+; \theta)$ with different learning strategies (**M**1 to **M**4) based on two criteria, including the reconstruction error and the log likelihood on the hold-out pairwise data. The reconstruction error refers to the average pixel difference between the transferred pseudo thermal eyes sampled from $p(\mathbf{t}^+|\mathbf{s}_j^+; \theta)$ and the ground truth thermal eyes. We calculate the log-likelihood on the hold-out testing set using the method in [17].

As can be seen in Table 2, while both the semi-supervised learning and the constraint (**M**2&**M**3) improve the performances over standard learning method **M**1, combining them together (**M**4, the full model) achieves the best performance.

Table 2. Evaluation of different constrained transfer learning methods in layer one.

| Methods | Reconstruction error | Log likelihood |
|---|---|---|
| M1: Pairwise data | 0.0627 | -514.1 |
| M2: Semi-supervised | 0.0601 | -484.8 |
| M3: Pairwise + constraint | 0.0588 | -492.0 |
| M4: Semi + constraint | **0.0580** | **-476.9** |

### 5.3.2 Evaluation of the thermal eye detector with the constrained deep transfer features

In this subsection, we further evaluate the eye detection performance based on the features learned using the proposed method with two-layer DBM (two iterations in Algorithm 1). For the constrained transfer learning step in each layer of the deep model, we use four different learning strategies (**M**1-**M**4).

**(1) Different methods:** We compare the proposed method with different strategies (**M**1-**M**4) to the baseline method (**M**0). The baseline method learns features using the standard Deep Boltzmann Machine model based on the thermal data without any knowledge transfer. As shown in Table 3, even with only the pairwise data, the proposed transfer feature learning method (**M**1) improves over the baseline (**M**0). The semi-supervised learning (**M**2) and constraint (**M**3) further boost the performances. By combining them together (**M**4), the constrained deep transferred features learned from multi-modality data increase the detection rate by 6.09%, comparing to the baseline (**M**0).

We also implement two other baselines using DBM and visible training data. (**V**0) refers to DBM feature learning based on the visible data. (**V**1) fine-tune the visible features learned with visible data on thermal data, so that the model parameters for thermal feature learning are initialized as the parameters of visible features. As can be seen in Table 3, they are all inferior to the proposed method (**M**4).

Table 3. Eye detection rates using different methods

| Methods | Eye detection rate |
| --- | --- |
| M0: DBM (baseline) | 87.45% |
| M1: Pairwise data | 89.48% |
| M2: Semi-supervised | 90.04% |
| M3: Pairwise + constraint | 91.88% |
| M4: Semi + constraint | **93.54%** |
| V0: DBM_visible (baseline) | 73.25% |
| V1: DBM_visible_finetune (baseline) | 86.72% |

**(2) Transfer in different layers:** Figure 4 shows that it's important to perform the constrained transfer learning in multiple layers, since it further boosts the performance comparing to one layer transfer. In addition, transferring in the deeper feature space (layer 2) is slightly better than transferring in the shallow feature space (layer 1).

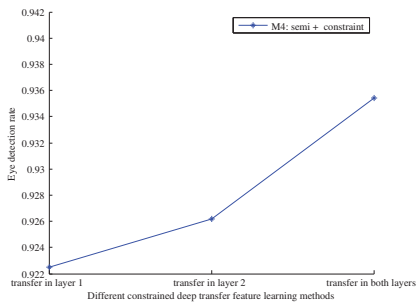

Figure 4. Transfer in different layers.

**(3) Learning with different amount of training data**
We vary the training data to train the proposed method (features and classifier) and the baseline method, and the eye detection rates are shown in Figure 5. Specifically, we keep reducing the number of training subjects from "P7+UP3" (7 subjects with pairwise data + 3 subjects with unpaired data) to "P2+UP2". Comparing **M**4 to **M**0, the proposed method always learns better features than the baseline. Comparing **M**4 to **M**2, the constraints are important and they improve the performances with different sets of data.

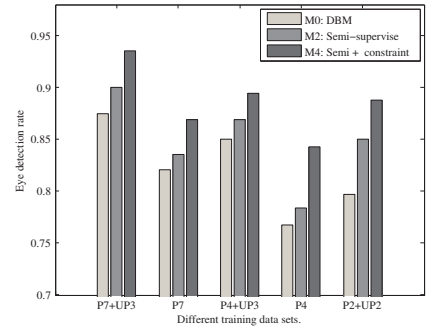

Figure 5. Learning with different amounts of training data.

**(4) Comparison with other features:** We compare the learned features with the proposed method (**M**4) to standard image features, such as the Scale Invariant Feature Transform (SIFT) feature, and the Histogram of Oriented Gradients (HOG) feature. For fair comparison, we train the linear SVM with same training data (P7+UP3), and the experimental results only differ due to the features. As shown in Figure 6, the learned features with the proposed method significantly outperforms the other designed features.

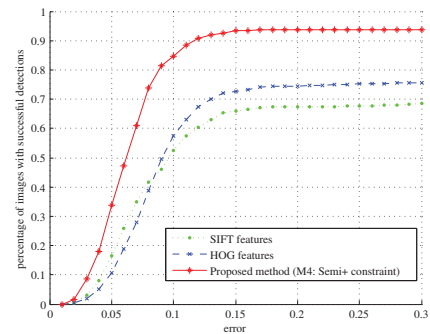

Figure 6. Cumulative error distribution curves using eye detectors with different features.

**(5) Visualization:** Figure 7 (a)(b) show the learned filters (parameters $\xi^1$ and $\xi^2$) with the proposed method (**M**4). Filters in the lower level capture local dependencies. They represent small dots, and some of them are similar to the Gabor Filters. Filters in the higher level capture more global variations.

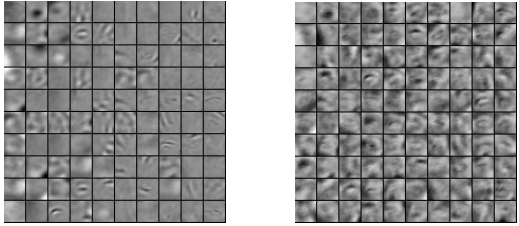

(a) learned filters in layer 1    (b) learned filters in layer 2

Figure 7. (a)(b): Learned filters in different layers.

### 5.4. Comparison with existing thermal eye detectors on other databases

To the best of our knowledge, there are only two works [10][22] that report eye detection results on publicly available thermal face databases. In [10], image patches are represented by Haar wavelet coefficients. Two successive classifiers (coarse and fine) are constructed based on features in different levels of details in a cascade manner. In [22], information from the whole face region is used to support eye detection on thermal facial images. It identifies 15 sub-regions on the face. Haar-like features and SVM are used to learn the eye detector.

Table 4 illustrates that our detector outperforms the method [10] on the MAHNOB Laughter database [13]. On the NVIE database [21], our detector is significantly better than the method in [22].

Table 4. Comparison with existing thermal eye detectors on the MAHNOB laughter database and the NVIE database.

| MAHNOB Laughter Database: | | |
|---|---|---|
| Methods | reported in [10] | Proposed method (**M4**) |
| Detection rate | 83.3% | 87.72% |
| **NVIE Database:** | | |
| Methods | reported in [22] | Proposed method (**M4**) |
| Detection rate | 68% | 81.60% |

### 5.5. Second Application

The proposed constrained deep transfer feature learning method is not limited to visible and thermal domains. It can be extended to other domains for simultaneous knowledge transfer and deep feature learning. In this section, we apply the proposed framework to cross-view facial expression recognition as the second application. The goal of this application is to learn the effective feature representation to classify neutral and non-neutral face in non-frontal view by transferring the data in frontal view (source) to non-frontal view (target). The unique facial structure in non-frontal view is used to constrain the transferring learning. In the experiments, we used the Multi-Pie database [6] (Figure 8), which contains facial images with varying head poses, facial expressions, and illumination conditions (training: id 1-200, testing: id 201-337).

Figure 9 below shows the results. In the experiments, we compared the classification accuracies (linear SVM) with different features, including the learned features using the proposed constrained deep transfer feature learning method, the learned features using the conventional DBM, the hand-crafted HOG and LBP features. In addition, we vary the number of training subjects in the target domain (100, 50, 20). There are a few observations. First, the learned features are significantly better than the hand-crafted features. Second, for all settings, by transferring the knowledge in the source domain, the learned features using the proposed method outperform the features learned with DBM.

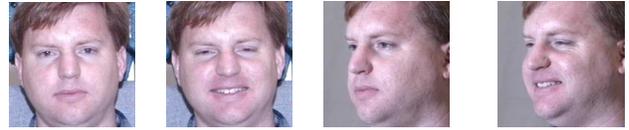

(a) Frontal image (source)    (b) Non-frontal image (target)

Figure 8. Cross-view neutral (left) and non-neutral (right) facial expression recognition.

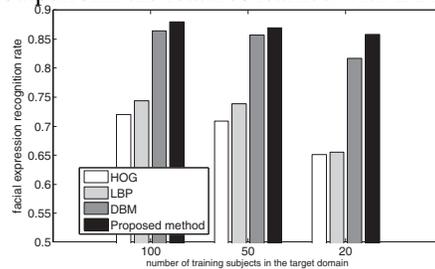

Figure 9. Facial expression classification accuracies with different features and numbers of training subjects in the target domain.

### 6. Conclusion

In this paper, we propose a constrained deep transfer feature learning framework in order to perform feature learning in a data-scarce target domain. Instead of performing transfer learning in a fixed feature space, we propose to simultaneously perform transfer and feature learning iteratively in increasingly higher level of feature spaces in order to minimize the semantic gap between the source and target domains. Furthermore, to ensure the transferred data to be semantically meaningful and to be consistent with the underlying properties of the target domain, we incorporate target domain knowledge as constraints into the transfer learning.

We applied the proposed framework for thermal feature learning for thermal eye detection by transferring the knowledge from visible domain. We also applied the it for cross-view facial expression recognition. The experiments demonstrate the effectiveness of the proposed framework for both applications. In the future, we would apply the framework to other vision applications.

**Acknowledgements:** The work described in this paper was supported in part by the National Science Foundation under the grant number NSF CNS-1205664 and in part by Army Research Office under the grant number ARO W911NF-12-C-0017.